%% file: main.tex
\documentclass[lettersize,journal]{IEEEtran}
\usepackage{amsmath,amsfonts}
\usepackage{algorithmic}
\usepackage{algorithm}
\usepackage{array}
\usepackage{multirow}
\usepackage[caption=false,font=normalsize,labelfont=sf,textfont=sf]{subfig}
\usepackage{textcomp}
\usepackage{stfloats}
\usepackage{url}
\usepackage{verbatim}
\usepackage{graphicx}
\usepackage{cite}
\usepackage{booktabs}
\usepackage{float}
\usepackage{color}
\makeatletter
\def\ps@IEEEtitlepagestyle{
  \def\@oddfoot{\mycopyrightnotice}
  \def\@evenfoot{}
}
\def\mycopyrightnotice{
  {\footnotesize
  \begin{minipage}{\textwidth}
  \centering
  Copyright~\copyright~2024 IEEE. Personal use of this material is permitted. However, permission to use this \\ material for any other purposes must be obtained from the IEEE by sending a request to pubs-permissions@ieee.org.
  \end{minipage}
  }
}

\begin{document}

\title{Learning Contrastive Self-Distillation for Ultra-Fine-Grained Visual Categorization \\Targeting Limited Samples}

\author{Ziye~Fang, Xin~Jiang, Hao~Tang, and Zechao~Li, \emph{Senior Member, IEEE}

\thanks{This work was partially supported by the National Natural Science Foundation of China (Grant No. U20B2064 and U21B2043). The first two authors contributed equally to this work. \emph{(Corresponding Author: Zechao Li.)}}

\thanks{Z. Fang, X. Jiang, H. Tang, and Z. Li are with the School of Computer Science and Engineering, Nanjing University of Science and Technology, Nanjing 210094, China  (e-mail: ziyefang@njust.edu.cn; xinjiang@njust.edu.cn;
tanghao0918@njust.edu.cn;
zechao.li@njust.edu.cn).
}
}

\markboth{IEEE Transactions on Circuits and Systems for Video Technology, 2024}%
{Shell \MakeLowercase{\textit{et al.}}: Bare Demo of IEEEtran.cls for IEEE Journals}


\maketitle

\input{sections/abstract.tex}

\input{sections/introduction.tex}
\input{sections/relatework.tex}

\input{sections/method.tex}

\input{sections/experiments.tex}
\input{sections/conclusion.tex}

\bibliographystyle{IEEEtran}
\bibliography{reference}

\end{document}

%% file: sections/abstract.tex
\begin{abstract}

    In the field of intelligent multimedia analysis, ultra-fine-grained visual categorization (Ultra-FGVC) plays a vital role in distinguishing intricate subcategories within broader categories. However, this task is inherently challenging due to the complex granularity of category subdivisions and the limited availability of data for each category.
    To address these challenges, this work proposes CSDNet, a pioneering framework that effectively explores contrastive learning and self-distillation to learn discriminative representations specifically designed for Ultra-FGVC tasks.
    CSDNet comprises three main modules: Subcategory-Specific Discrepancy Parsing (SSDP), Dynamic Discrepancy Learning (DDL), and Subcategory-Specific Discrepancy Transfer (SSDT), which collectively enhance the generalization of deep models across instance, feature, and logit prediction levels.
    To increase the diversity of training samples, the SSDP module introduces adaptive augmented samples to spotlight subcategory-specific discrepancies.
    Simultaneously, the proposed DDL module stores historical intermediate features by a dynamic memory queue, which optimizes the feature learning space through iterative contrastive learning.
    Furthermore, the SSDT module effectively distills subcategory-specific discrepancies knowledge from the inherent structure of limited training data using a self-distillation paradigm at the logit prediction level.
    Experimental results demonstrate that CSDNet outperforms current state-of-the-art Ultra-FGVC methods, emphasizing its powerful efficacy and adaptability in addressing Ultra-FGVC tasks.

\end{abstract}


%% file: sections/introduction.tex
\begin{figure}
  \centering
  \begin{tabular}{c}
  \includegraphics[width=0.98\linewidth]{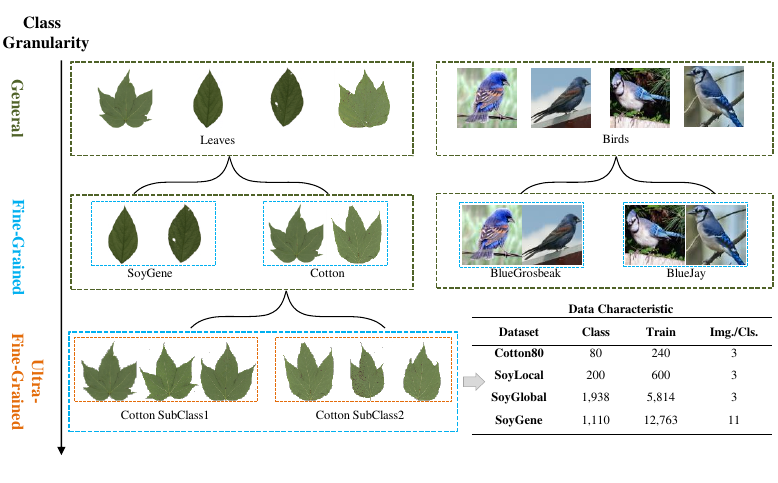} 
  \end{tabular}
  \caption{An illustration of the differences among Ultra-FGVC, FGVC, and general categorization, and the data size available for each class in Ultra-FGVC. Img./Cls. denotes the number of samples in each category. As the granularity of data categorization becomes increasingly fine, the number of available samples per class in the dataset becomes very limited, with an average of only 3 to 11 images per class.}
  \label{fig:ultra}
\end{figure}

\section{Introduction}

\IEEEPARstart {U}{ltra-Fine-Grained} Visual Categorization (Ultra-FGVC) is dedicated to distinguishing subcategories within the same fine-grained category. 
It has significant applications in intelligent multimedia analysis, such as biological species identification (\emph{e.g.,} birds and plants) and crop pests and diseases categorization~\cite{c2,c57,c61}.
With the advancement of precision agriculture, achieving a more detailed fine-grained categorization has become increasingly important.
However, Ultra-FGVC is further complicated by the presence of both high inter-class similarity and substantial intra-class variations, similar to FGVC tasks~\cite{c58,c66,ZhaTST23}. Additionally, it faces challenges from the scarcity of data samples due to decreased categorization granularity, as illustrated in Figure~\ref{fig:ultra}.

With the advancements in deep convolutional neural networks (CNNs)~\cite{c35, c62}, significant progress has been achieved in the field of Ultra-FGVC, which yields promising results~\cite{c9, c5, c1}. CNNs are naturally suitable for discovering discriminative regions in images. However, it still presents a significant challenge \cite{c63} to understand the intricate interrelations among these regions. It is crucial to understand subtle local details and complex global content to effectively differentiate subcategories in the Ultra-FGVC task. In this regard, vision transformers~\cite{c22} have emerged as powerful feature extractors due to their potent self-attention mechanisms and ability to process image tokens. These attributes enable transformers to capture both local details and global contextual information~\cite{c59, c60} and make them highly suitable for Ultra-FGVC tasks. 
Recent studies~\cite{c3, c6} have employed transformers for Ultra-FGVC tasks, which demonstrate significant advantages over traditional CNNs-based methods~\cite{c1,c5,c9}. These results validate the revolutionary power and efficacy of the transformer architecture in Ultra-FGVC. Nevertheless, the integration of large-scale deep learning models in Ultra-FGVC still encounters challenges that require further exploration and deliberate consideration to fully exploit their potential in this intricate task.
\IEEEpubidadjcol

A primary challenge in Ultra-FGVC is how to guide large-scale deep models to learn an optimized feature space under limited training samples, which should maximize inter-class distance and maintain tolerance to intra-class variations at the same time.
Current methods address this challenge through different data augmentation strategies, commonly in the form of patch-level~\cite{c9,TangLPT20} and part-level~\cite{c5,c3} augmentations.
Patch-level augmentation subdivides images into uniform square patches to extract patch-level feature descriptions. And part-level augmentation generates augmented images by randomly erasing select image regions. However, these strategies compromise semantic integrity, which leads to a deficiency in discriminative granularity required for precise categorization. 
Besides, we observe that existing instance-level approaches~\cite{c94,c6,TangYLT22} primarily focus on enhancing the discriminative representation through carefully designed data augmentation loss functions. 
Therefore a question naturally arises: is it possible that we can design an instance-level data augmentation method to enhance subtle discrepancies specific to each subclass instance?
We also observe that existing works tend to overlook the potential present in intermediate features and logit prediction elements that reflect different semantic interpretations of the input instances by the model.
Therefore, how to exploit the potential of these neglected dimensions, focusing on the effective use of intermediate features and logit predictions is worthy of investigation.

Inspired by the above observations, we propose a novel Contrastive Self-Distillation framework (CSDNet) that addresses Ultra-FGVC issues at three key levels: instance, feature, and logit prediction. This approach aims to mitigate the limitations caused by limited sample availability.
At the instance level, we propose the Subcategory-Specific Discrepancy Parsing (SSDP) module. This module identifies specific discrepancies within subcategories by parsing semantic features and generates augmented samples. These samples provide enhanced discriminatory insights because they amplify subtle discrepancies, reduce background noise, and eliminate non-essential information.
For feature-level refinement, the Dynamic Discrepancy Learning (DDL) module containing a dynamic memory queue is proposed. This module aims to incorporate intermediate features from historical training samples and integrate them within a contrastive iteration using contrastive learning. This process furnishes CSDNet with an optimized feature space, which increases its sensitivity to specific discrepancies across various subcategories. 
However, an issue that cannot be overlooked is that utilizing augmented samples for prediction leads to a highly time-consuming and memory-demanding evaluation process. Thereby, we introduce the Subcategory-Specific Discrepancies Transfer (SSDT) module at the logit prediction level.
The SSDT module adopts a self-distillation paradigm, effectively extracting subcategory-specific discrepancies knowledge from the inherent structure of limited training data, enabling CSDNet to make predictions using only raw samples.
%

%
The main contributions of this paper can be summarized as follows:
\begin{itemize}
\item{
This work proposes a CSDNet framework that leverages the advantages of contrastive learning and self-distillation to alleviate the limited sample problem in Ultra-FGVC tasks.
}
\item{The proposed Subcategory-Specific Discrepancy Parsing and Dynamic Discrepancy Learning methods expand training samples at both the instance and feature levels. Meanwhile, the Subcategory-Specific Discrepancy Transfer method learns subcategory-specific discrepancy knowledge from limited samples through a self-distillation mechanism at the logit prediction level.
}
\item{CSDNet has been evaluated on five ultra-fine-grained and four fine-grained visual classification benchmark datasets. Experimental results demonstrate that CSDNet establishes a new state-of-the-art performance.}
\end{itemize}

The rest of this paper is organized as follows. Section~\ref{sec:two} provides a brief review of related work. Section~\ref{sec:three} explains on the proposed method, including the model architecture and loss functions. Section~\ref{sec:four} covers experimental details, results, and related analyses. Finally, in section~\ref{sec:five}, the conclusion is presented.

%% file: sections/relatework.tex
\IEEEpubidadjcol
\section{Related Works}\label{sec:two}
In this section, we explore recent methods related to this work, focusing on three primary directions: ultra-fine-grained visual categorization, contrastive learning, and knowledge distillation.

\subsection{Ultra-fine-grained Visual Categorization} 

Ultra-fine-grained visual categorization (Ultra-FGVC)\cite{c1,c5,c9} aims to differentiate subcategories with very fine granularity, often so similar that even the human eye struggles to distinguish them. This area has received a lot of attention\cite{c6,c86,c87} in recent years.
Compared with fine-grained visual categorization (FGVC), Ultra-FGVC is labeled by genetic databases\cite{c8}, instead of relying on experts or volunteers. 
The previous study\cite{c90} introduced the first ultra-fine-grained dataset consisting of 422 leaf images and represented only 3 cultivars of the same species. Unfortunately, this dataset was not accessible to the public.
Recently, MORT\cite{c1} proposed a public ultra-fine-grained dataset with 600 images from 100 soybean varieties. 
Subsequently, a large Ultra-FGVC dataset named UFG\cite{c8} was introduced, which contained 47,114 leaf images from 3,526 subcategories, and established a benchmark platform with baseline performances on 13 CNN techniques.
Since the Ultra-FGVC task has a large number of classification subcategories and the total number of samples is limited, it is distinct from FGVC primarily due to its extremely limited data constraints. This makes Ultra-FGVC more challenging than FGVC.
Current methods mainly solve this challenge from two levels of data augmentation: patch-level and part-level.

The patch-level data augmentation divides images into a set of square patches, thereby providing the patch-level feature descriptions.
For example, MaskCOV\cite{c9} is a CNN-based random mask covariance network, which shuffles and masks image patches and then predicts the original position of each patch through a self-supervised learning module. 
Similarly, the recent CLE-ViT\cite{c6} generates augmented images by randomly masking and shuffling patches.
The part-level data augmentation aims to obtain semantically meaningful parts and generates augmented images by erasing or mixing these parts.
For instance, SPARE\cite{c5} employs solely subcategory labels in an unsupervised manner for part segmentation and erasing and predicts the contextual location of the erased part during training.
In Mix-ViT\cite{c3}, an unsupervised learning module is designed, which mixes partial features and learns to predict whether a partial feature has been replaced. 

Based on their studies, we recognize that Ultra-FGVC tends to result in overfitting due to constraints in data samples and suboptimal feature extraction capabilities. 
Previous works~\cite{c3,c5,c9} are mainly based on patch-level or part-level data augmentation to tackle the challenge. However, operations such as erasing these data augmentation methods might compromise the integrity of the ultra-fine-grained image semantics, subsequently degrading classification accuracy. 
In this paper, instance-level data augmentation is introduced, which preserves complete semantic information while generating multiple view images containing subcategory-specific discrepancies, laying the foundation for subsequent modules.

\subsection{Contrastive Learning} 
Contrastive learning (CL), as demonstrated by a multitude of studies \cite{c76}, has achieved remarkable success across various tasks\cite{c80, KSTN2023}, especially in the context of unsupervised learning.
%
The model is trained pairwise samples constructed through various data augmentations. It evaluates the similarities between these samples and aims to maximize the likeness of positive pairs and minimize the likeness of negative pairs.
The objective of contrastive learning is to enhance the generalizability of the representations and contribute to a range of subsequent tasks. 
Recently, many studies have introduced it in the form of a loss function into supervised tasks\cite{c6, c28}, yielding promising results. 
Supervised contrastive learning (SupCon)\cite{c85} extends contrastive learning into the realm of fully-supervised tasks. It pairs samples of the same class as positive pairs and samples from different classes as negative pairs. This approach outperforms traditional supervised cross-entropy loss in the context of image categorization.

Current supervised learning primarily focuses on mining information from the current features. However, the exploration of historical information has largely been ignored. 
Some studies\cite{c24,c7} in unsupervised learning, leverage historical information to optimize the feature space. For instance, MoCo\cite{c7} constructs a dynamic dictionary using a queue combined with a moving-average encoder. 
It has shown that features from historical batches positively influence model convergence, suggesting that valuable knowledge can be extracted from previous features. 
In this paper, we introduced the queue contrastive loss to Ultra-FGVC, which employs a queue to retain intermediate features of historical training samples and expand samples at the feature level, using the latent knowledge from historical features during the learning of current ones to be more sensitive to subcategory-specific discrepancies.

\subsection{Knowledge Distillation} 

Knowledge distillation (KD)\cite{c23} utilizes knowledge from a complex pre-trained model (i.e., teacher network) to guide the learning of a smaller, lightweight network (i.e., student network) via soft labels. 
As the research progressed, various strategies for knowledge distillation were investigated. Notably, there has been a substantial exploration into the concept of leveraging the student network as its own instructor, a technique referred to as ``self-knowledge distillation" (self-KD)\cite{c14}.
It means that a student model becomes a teacher model itself. One method of self-KD is to reduce the feature distance between similar inputs of a model \cite{c15}, arising from data augmentation of the same sample or from different samples within the same class. 
For example, DDSD \cite{c16} introduced a knowledge transfer framework using image distortion, generating two distorted images through random mirroring and cropping. Similarly, CS-KD \cite{c14} employs two different samples from the same class. Both DDSD and CS-KD aim to minimize the distance between features extracted from these images. 
The other approach is to directly use outputs from a teacher whose architecture is exactly the same as a student \cite{c15}. 
The Born-Again Network (BAN) \cite{c17} initially trains a network, and then employs this pre-trained model as the teacher for its successor. This procedure is iteratively executed, culminating in multi-generational KD. Then, the knowledge from the ($k$-1)-th model is utilized to train the $k$-th generation model. 

Self-KD has been investigated within, contrastive learning\cite{c19}, semi-supervised learning\cite{c20} and self-supervised learning\cite{c21}. 
This paper argues that there is a knowledge disparity between augmented images and raw ones, with the former containing more subcategory-specific discrepancies. 
Therefore, self-distillation is introduced to extract knowledge of subclass-specific discrepancies in logit predictions, which can provide stronger feedback to the model, especially in low-data situations\cite{c12}.

%% file: sections/method.tex
\begin{figure*}
  \centering
\includegraphics[width=0.98\textwidth]{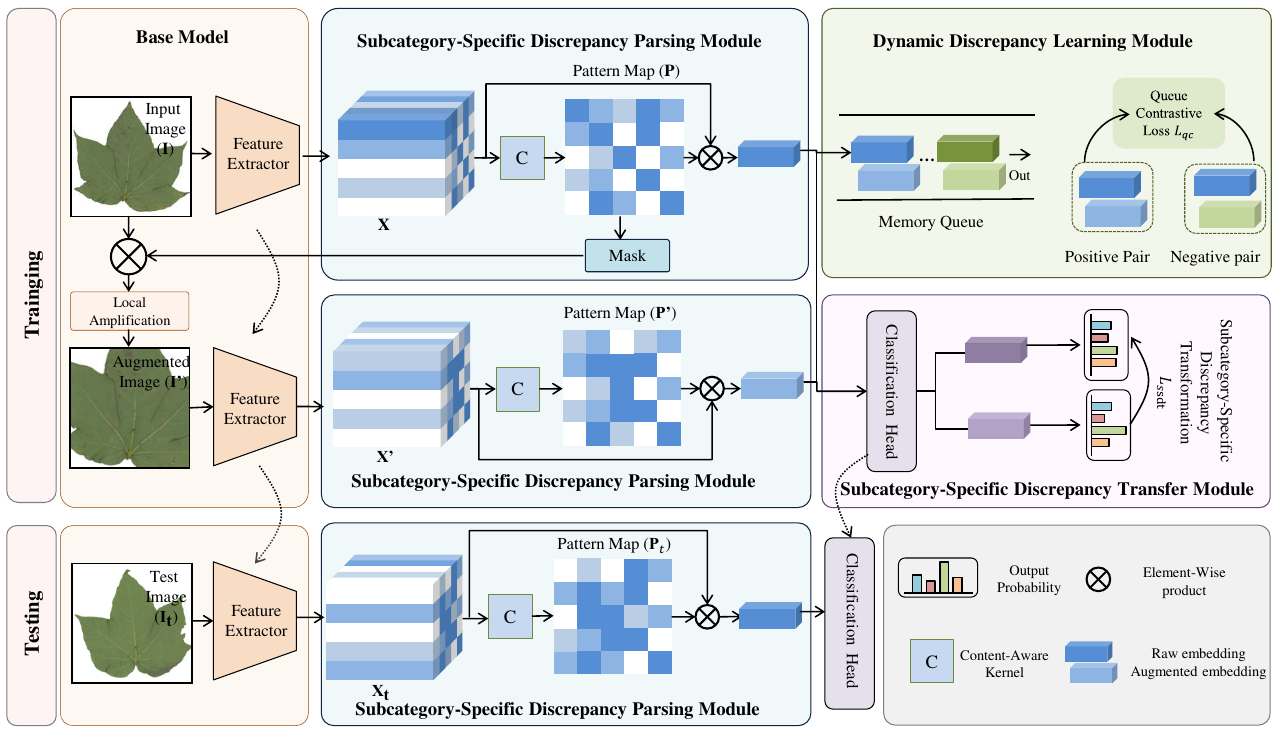} 
  \caption{An overview of the proposed method. Firstly, a feature $\mathbf X$ is extracted from the input image $\mathbf I$ using the backbone. Secondly, the Subcategory-Specific Discrepancy Parsing (SSDP) module uses a content-aware kernel to obtain a pattern map $\mathbf P$. Using this map, a subcategory-specific discrepancy mask is created, which generates the augmented image $\mathbf I'$. This augmented image is then sent to the backbone for retraining. The pattern map $\mathbf P$, after activation by the sigmoid function, is applied to the feature $\mathbf X$ to enhance the representation of subcategory-specific discrepancies. Thirdly, historical image features are stored in a memory queue and these current features are then integrated using queue contrastive learning. Lastly, self-distillation is leveraged to distill subcategory-specific knowledge between raw and augmented images.}
  \label{fig:framework}
\end{figure*}
\section{Method}\label{sec:three}
\subsection{Overview of CSDNet}

Given the limited samples available in Ultra-FGVC, we argue that expanding the training dataset efficiently can address this challenge.
To this end, we introduce the Subcategory-Specific Discrepancy Parsing (SSDP) module and the Dynamic Discrepancy Learning (DDL) module. The former aims to amplify subcategory-specific discrepancies and generate augmented samples. The latter expands training samples at the feature level by introducing memory queues to store intermediate features of historical training samples for contrastive learning.
In addition, the Subcategory-Specific Discrepancy Transfer (SSDT) module is responsible for learning more subcategory-specific discrepancy knowledge from the limited samples through a distillation mechanism.
Notably, the DDL module and the SSDT module are only conducted during training to enhance the generalization capacity of the feature extractor.

\subsection{Subcategory-Specific Discrepancy Parsing} \label{sec3.c}

The main challenge confronting Ultra-FGVC at present is the extremely limited samples. Recent works try to address this issue by augmenting data at both the patch-level~\cite{c3} and part-level~\cite{c5,c9}. 
However, certain operations in these methods, such as erasing, may compromise crucial semantic information necessary for accurate categorization. Therefore, we introduce the Subcategory-Specific Discrepancy Parsing (SSDP) module, which contains an instance-level data augmentation method focused on generating augmented images that are semantically continuous and contain subcategory-specific discrepancies.
Specifically, the SSDP module is composed of subcategory-specific discrepancy parsing and subcategory-specific discrepancy refinement.
The former generates augmented samples that provide augmented discriminative insights by parsing the semantic features extracted by the feature extractor. The latter allows us to better focus on crucial features via amplifying subtle discrepancies.

\begin{figure}
  \centering
  \begin{tabular}{c}
  \includegraphics[width=0.9\linewidth]{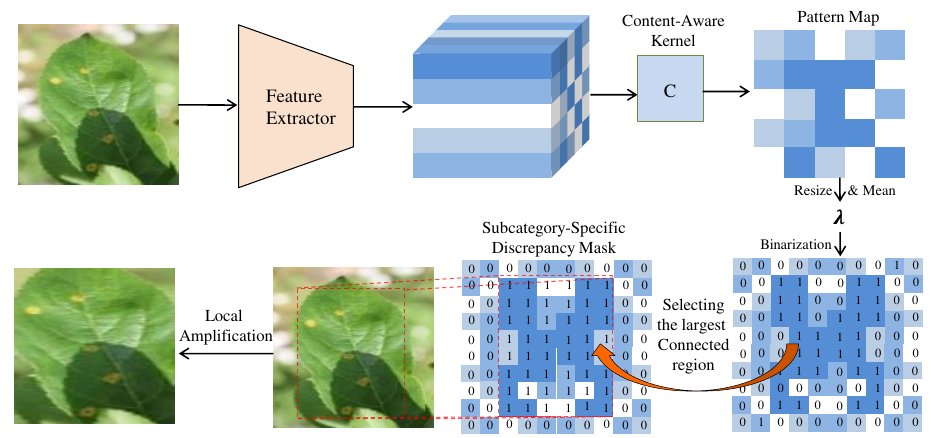} 
  \end{tabular}
  \caption{The implementation details of generating augmented images with the SSDP module. 
  The binarization of the image is implemented by Eq. (\ref{eq:1}).}
  \label{fig:ssdpdetail}
\end{figure}

\textbf{Subcategory-specific discrepancy parsing.} 
To obtain subcategory-specific discrepancies as shown in  Fig. \ref{fig:ssdpdetail}, we utilize a content-aware kernel to transform the high-level semantic features into low-level details, which specify the location, scale, and intensity of the discrepancies.
Concretely, given the image $\mathbf I \in \mathbb{\mathbb{R}}^{3\times {I_h}\times {I_w}}$, we feed $\mathbf I$ into feature extractor to generate the high-level semantic feature $\mathbf X \in \mathbb{R}^{C\times H\times W}$. 
Next, we transform the $\mathbf X$ into a pattern map $\mathbf P \in \mathbb{R}^{H\times W}$ through a content-aware kernel $\mathbf W_K \in \mathbb{R}^{C\times 1\times 1}$. The calculation formulation is shown below:
\begin{equation}
\mathbf P(x,y) = \sum_i^C{\mathbf W_k(i)\cdot \mathbf X(i,x,y)},
\label{eq:P}
\end{equation}
where $\mathbf W_k(i)$ represents the value of the content-aware kernel at channel $i$ and $\mathbf X(i,x,y)$ represents the value of the high-level features at channel $i$ and spatial location $(x, y)$. 
Then we resize the pattern map to the input image, denoted as $\mathbf {\hat{P}}$ that is binarized using Eq. (\ref{eq:1}). Next, we select the largest connected area on $\mathbf {\hat{P}}$, thereby obtaining a complete region mask called subcategory-specific discrepancy mask $\mathbf M \in \mathbb{R}^{{I_h}\times {I_w}}$ from the discrete binarized map:
\begin{align}
\label{eq:1}
\mathbf B(i,j) &=
\left\{
\begin{aligned}
   &  1,\; ~~\textrm{if}\;  \mathbf {\hat{P}}(i,j) > \lambda,\lambda=\textrm{mean}(\mathbf {\hat{P}})\\ \\
   &  0,\; ~~\textrm{otherwise}
\end{aligned},
\right.\\
\label{eq:mask}
\mathbf M(i,j) &= \textrm{op}(\mathbf B(i,j)), 
\end{align}
where $\textrm {mean}(\cdot)$ represents the computation of the average value for the elements in $\mathbf {\hat{P}}$. $\textrm {op}(\cdot)$ means selecting the largest connected region from binarized image $\mathbf B(i,j)$. This operation sets the mask values within the largest connected area to 1, while all other mask values are set to 0, creating a continuous square area.


With the subcategory-specific discrepancy mask $\mathbf M$, we can obtain the augmented image that contains subcategory-specific discrepancies rich in semantic information and intricate details from the raw image, as follows:
\begin{equation}
\mathbf{ I'} = \mathbf I \odot \mathbf M,
\label{eq:I'}
\end{equation}
where $\odot$ denotes element-wise multiplication. It is worth noting that these highly responsive regions that are preserved by the mask are scaled up to the original image size ${3\times {I_h}\times {I_w}}$.
The augmented image $\mathbf{I'}$ is fed back into the feature extractor for further training.

\textbf{Subcategory-specific discrepancy refinement}. 
While subcategory-specific discrepancy parsing generates augmented images that provide augmented discriminatory insight, it can not directly help the model refine its focus on crucial features.
Therefore, we use the pattern map $\mathbf P$ to spotlight subcategory-specific discrepancies and reduce background noise and non-essential information, thereby allowing the model to refine focus on discrepancy regions:
\begin{equation}
\mathbf X_s = \textrm{sigmoid}{(\mathbf P)} \odot \mathbf X,
\label{eq:xx}
\end{equation}
where $\odot$ means element-wise product. 

The SSDP module serves three primary purposes. 
First, it provides a wider array of training samples, which enhances the ability of the model to tolerate intra-class variations. 
Second, the diversity of samples augments the inter-class distance, facilitating the differentiation of highly similar samples across different subcategories. 
Finally, the abundance of samples and the refinement of subcategory-specific discrepancies can improve the representational capacity of features when working with limited samples.
During training, subcategory-specific discrepancy parsing is used to expand the limited training samples and enable the model to acquire more discrepancies knowledge. However, during testing, only the raw image undergoes discrepancy refinement after feature extraction, which means the SSDP module does not produce augmented images.
Consequently, this significantly reduces the inference time complexity. Moreover, as both raw and augmented images use the same feature extractor for feature extraction, additional parameters are not introduced.

\subsection{Dynamic Discrepancy Learning} \label{sec3.d}

In unsupervised learning, techniques like Memory Bank\cite{c24} and MoCo\cite{c7} show the value of allocating space for historical features and expanding the dictionary used for contrastive learning.
Their successes underline the importance of leveraging historical feature information. However, in supervised learning, the emphasis has been predominantly on current features, often ignoring the rich potential of historical data.
Based on the above observation, our aim is to integrate historical data into supervised contrastive learning. To achieve this, we introduce the Dynamic Discrepancy Learning (DDL) module, which utilizes a dynamic memory queue to store intermediate features from historical training samples.
During a single contrastive feature learning iteration, we integrate these stored features with the current features to calculate the contrastive loss.
This historical data provides the CSDNet with an optimized feature space, thereby increasing its sensitivity to subcategory-specific discrepancies.

By feeding the raw images and augmented images to the feature extractor, we can obtain the corresponding raw visual embeddings $\mathbf E_{raw}$ and augmented visual embeddings $\mathbf E_{aug}$. Then, the embeddings of the $m$-th batch can be expressed as:
\begin{equation}
\begin{split}
\mathbf E^{m} = \{ \mathbf E_{aug_a}^{m}, \mathbf E_{raw_a}^{m}, \dots, \mathbf E_{aug_b}^{m}, \mathbf E_{raw_b}^{m}\},
\end{split}
\end{equation}
and the memory queue can be expressed as:
\begin{equation}
\begin{split}
\mathbf M_{queue} = \{ \mathbf E_{aug_a}^{m-\iota}, \mathbf E_{raw_a}^{m-\iota}, \dots, \mathbf E_{aug_b}^{m-\iota}, \mathbf E_{raw_b}^{m-\iota}, \\ \mathbf E_{aug_a}^{m-\iota+1}, \mathbf E_{raw_a}^{m-\iota+1}, \dots, \mathbf E_{aug_{c}}^{m-1}, \mathbf E_{raw_{c}}^{m-1}\},
\end{split}
\label{mqueue}
\end{equation}
where $a, b, c$ are the subcategory label and $m$ represents the $m$-th batch. Notably, the length of $\mathbf M_{queue}$ is $\iota$, which means that the queue can store $\iota$ batch of data. 
We then aggregate historical and current features via concatenation (denoted as Cat):
\begin{equation}
\begin{split}
\mathbf M = \textrm{Cat} (\mathbf E^{m}, \mathbf M_{queue}),
\end{split}
\end{equation}

\textbf{Positive Pair Construction.}  
To better learn subcategory-specific discrepancies knowledge, we consider both raw images and augmented images in the positive pair construction.
In detail, the anchor images $\mathbf I$ are taken from the dataset, and the augmented images $\mathbf I'$ produced by the SSDP module serve as a positive sample. 
If their visual embeddings originate from the same subcategory, they are considered a positive pair~\cite{c27}:
\begin{equation}
P_{pos} =
\left\{
\begin{aligned}
   &  < \mathbf E_{raw_a}^{i}, \mathbf E_{aug_a}^i>\\
   &  <\mathbf E_{raw_a}^{j}, \mathbf E_{aug_a}^i>\\
   &  <\mathbf E_{raw_a}^{j}, \mathbf E_{raw_a}^i>\\
\end{aligned}
\right.
i,j=m-\iota,m-\iota+1,...,m. 
\label{pos}
\end{equation}

\textbf{Negative Pair Construction.} Negative sample pairs consist of sample pairs from different subcategories.
Specifically, for each image, we randomly select an image (excluding itself) from $\mathbf M$ as its negative sample, forming a negative pair:
\begin{equation}
P_{neg} =
\left\{
\begin{aligned}
   &  <\mathbf E_{aug_a}^{i}, \mathbf E_{raw_b}^i>\\
   &  <\mathbf E_{aug_a}^{j}, \mathbf E_{aug_b}^i>\\
   &  <\mathbf E_{raw_a}^{j}, \mathbf E_{raw_b}^i>\\
\end{aligned}
\right.
i,j=m-\iota,m-\iota+1,...,m. 
\label{neg}
\end{equation}

\textbf{Update Memory Queue $\mathbf M_{queue}$.} 
In contrastive learning, a larger dictionary often represents better learning performance.
However, with the iteration of training rounds, there is an accumulation of excessive feature differences,  which will harm contrastive learning.
As a result, we employed the ``First In First Out" property of the queue, which allows for the gradual replacement of samples within the queue:
\begin{align}
\mathbf M_{queue} &= \texttt{Dequeue}(\mathbf M_{queue}),\\
\mathbf M_{queue} &= \texttt{Enqueue}(\mathbf M_{queue},\mathbf E^{m}),
\end{align}
where $\texttt{Dequeue}$ means dequeue the earliest batch and $\texttt{Enqueue}$ denotes enqueue all current batch embeddings $\mathbf E^{m}$.

\textbf{Dynamic Discrepancy Learning.} We employ a contrastive loss $\mathcal L_{qc}$ to supervise the learning of positive $P_{pos}$ and negative sample pairs $P_{neg}$. 
It minimizes the similarity between $P_{pos}$ and maximizes the similarity between $P_{neg}$. To prevent the loss from being dominated by simple negative samples (from distinct classes with little similarity), a constant margin $\xi$ is introduced\cite{c28}, such that only negative sample pairs with a similarity greater than the margin contribute to the loss $\mathcal L_{qc}$. 
Formally, the contrastive loss is denoted as:
\begin{equation}
\begin{split}
\mathcal L_{qc} = \frac{1}{Q^2}\sum_u^Q[\sum_v^Q(1-\textrm{sim}(P_{pos}(u,v)))+ \\ \sum_v^Q\max{(\textrm{sim}(P_{neg}(u,v))-\xi,0)}],
\label{q12}
\end{split}
\end{equation}
where $Q$ is the amount of visual embedding in the $\mathbf M$. $P_{pos}$ and $P_{neg}$ are pre-processed with $L_2$ normalization and $\textrm{sim}(\cdot)$ means dot product.

\subsection{Subcategory-Specific Discrepancy Transfer} \label{sec3.e}
The SSDP and DDL modules effectively address the challenges posed by limited data at both the instance and feature levels, aiding the model in learning subcategory-specific discrepancy knowledge.
Naturally, we believe that there is untapped knowledge about subcategory-specific discrepancies in the logit predictions of raw and augmented samples.
Meanwhile, a non-negligible problem is that utilizing augmented samples for prediction leads to very time-consuming and memory-demanding during evaluation.

Network distillation~\cite{c23} is proven to be a viable solution for addressing the problems mentioned above.
Motivated by this,  we introduce the Subcategory-Specific Discrepancy Transfer (SSDT) module. This module aims to further distill subcategory-specific discrepancy knowledge through a self-distillation process between raw and augmented samples.

Formally, the raw sample logit prediction $\mathbf y_{raw}$ and the augmented sample logit prediction $\mathbf y_{aug}$ are used for self-distillation:
\begin{align}
\mathbf y_{raw} &= \textrm{Head}(\mathbf E_{raw}), ~\mathbf y_{aug} =\textrm{Head}(\mathbf E_{aug}),\\
\mathcal L_{ssdt} &=\textrm{KL}(\mathbf y_{raw}, ~\mathbf y_{aug}),
\end{align}
where $\textrm{Head}(\cdot)$ represents a classification head, composed of a linear layer. $\textrm{KL}(\cdot, \cdot)$ denotes a Kullback Leibler divergence loss.

In contrast to $\mathbf y_{raw}$ that includes some noise and background information, $\mathbf y_{aug}$ contains more subcategory-specific discrepancies targeting ultra-fine-grained knowledge. 
Therefore, when the model is optimizing for $\mathcal L_{ssdt}$, it distills more subcategory-specific discrepancies knowledge from limited samples without requiring additional annotation. This enhances model generalization and mitigates overfitting caused by limited samples.
Furthermore, we are able to make inference predictions only using $\mathbf y_{raw}$, thus reducing the time required for inference. 

\subsection{Loss Function}
In summary, the overall training loss for the proposed CSDNet can be formulated as: 
\begin{equation}
\mathcal L = \mathcal L_{cls} + \alpha \mathcal L_{qc} + \beta \mathcal L_{ssdt},
\label{eq:l}
\end{equation}
where $\mathcal L_{cls}$ denotes Label Smoothing CrossEntropy loss function~\cite{c49}, $\alpha$ and $\beta$ are the hyper-parameters to weight the $\mathcal L_{qc}$ loss and $\mathcal L_{ssdt}$ loss.

%% file: sections/experiments.tex
\section{Experiments}\label{sec:four}
\begin{table}[t]
    \centering
    \caption{Statistics of the benchmark datasets.}
    \begin{tabular}{cccccc}
        \toprule
        Granularity & Dataset & \#Class & \#Train & \#Test \\
        \midrule
        \multirow{5}{*}{Ultra-fine-grained}& Cotton80 & 80 & 240 & 240 \\
        & SoyLocal & 200 & 600 & 600 \\
        & SoyGene & 1,110 & 12,763 & 11,143 \\
        & SoyAgeing & 198 & 4,950 & 4,950 \\
        & SoyGlobal & 1,938 & 5,814 & 5,814 \\
        \midrule
        \multirow{4}{*}{Fine-grained}& CUB-200-2011 & 200 & 5,994 & 5,794 \\
        & Food101 & 101 & 75,750 & 25,250 \\
         & IP102 & 102 & 45,095 & 22,619 \\
        & Apple Foliar disease & 4 & 1,366 & 455 \\
        \bottomrule
    \end{tabular}
    \label{tab:table2}
\end{table}
\subsection{Experimental Setup}

\subsubsection{Dataset}
We evaluate CSDNet on nine datasets, including five ultra-fine-grained image datasets \cite{c1}, Cotton80, SoyLocal, SoyGene, SoyAgeing, SoyGlobal, and four fine-grained image datasets Apple Foliar disease (AFD) dataset\cite{c29}, IP102\cite{c31}, Food101\cite{c97}and CUB-200-2011 (CUB) \cite{c30}. 
Cotton80, SoyLocal, SoyGene, and SoyGlobal have limited samples per subcategory, approximately 3 to 11 images. 
AFD is a dataset of crop diseases. IP102 is a long-tail dataset of crop pests with substantial intra-class variations, and high inter-class similarity. Table~\ref{tab:table2} shows the details of the nine datasets.

\subsubsection{Implementation Details}
In our experiments, we employ the Swin transformer base\cite{c22} as our backbone network and initialize it with pre-trained parameters on ImageNet21k\cite{c32}. 
For all datasets, we resize the images to $512\times 512$ and then center crop them to $448\times 448$ during both the training and inference phases. 
During training, we utilize the AdamW optimizer and batch size of 12 for all datasets.
For all datasets, the initial learning rate is set to 1e-3, except for SoyGene where it is set to 2e-3.
In Eq. (\ref{eq:l}), $\alpha$ is set to 1 and $\beta$ is set to 0.4.
For the memory queue size $\iota$, it is set to 1 for SoyGene and SoyGlobal, and 2 for the other datasets.
The margin $\xi$ in Eq. (\ref{q12}) is 1. For all experiments, we adopt top-1 accuracy as the evaluation metric.

\begin{table*}[htb]
    \centering
    \caption{The classification accuracies on the benchmark datasets. The results of the best-performing method are in boldface. Here A.F. represents apple foliar disease. L and FT indicate linear and fine-tuning evaluation, respectively. ViT-B means visual transformer Base and swin-B means Swin transformer base.}
    \begin{tabular}{cccccccc}
        \toprule
          \multirow{2}{*}{Method} & \multirow{2}{*}{Backbone} & \multicolumn{6}{c}{Accuracy (\%)} \\
        \cmidrule(l){3-8}
        &  & Cotton80 & SoyLocal & SoyGene & SoyAgeing &  SoyGlobal & A.F. \\
        \midrule
        Alexnet\cite{c33}&Alexnet&22.92&19.50&13.12&44.93&13.21&95.16\\
        VGG-16\cite{c34}&VGG-16&39.33&39.33&63.54&70.44&45.17&95.60\\
        ResNet-50\cite{c35}&ResNet-50&52.50&38.83&70.21&67.15&25.59&94.73\\
        \midrule
        SimCLR (FT\cite{c25})&ResNet-50&51.67&37.33&62.68&64.73&42.54&93.63\\
        SimCLR (L)\cite{c25}&ResNet-50&41.25&29.17&29.62&46.18&13.48&82.86\\
        MoCo v2\cite{c37}&ResNet-50&45.00&32.67&56.49&59.13&29.26&96.04\\
        MoCo v2 (L)\cite{c37}&ResNet-50&30.42&27.67&26.58&38.26&12.99&85.49\\
        BYOL (FT)\cite{c38}&ResNet-50&52.92&33.17&60.65&64.75&41.35&96.04\\
        BYOL (L)\cite{c38}&ResNet-50&47.92&25.50&35.13&49.53&18.44&87.03\\
        \midrule
        Cutout (16)\cite{c39}&ResNet-50&55.83&37.67&61.12&65.70&47.06&94.95\\
        Cutout (8)\cite{c39}&ResNet-50&54.58&31.67&62.46&63.68&44.65&94.95\\
        ADL (0.25)\cite{c41}&ResNet-50&43.75&34.67&55.19&61.70&39.35&96.04\\
        ADL (0.5)\cite{c41}&ResNet-50&40.83&28.00&52.18&51.56&29.50&94.51\\
        Cutmix\cite{c42}&ResNet-50&45.00&26.33&66.39&62.68&30.31&93.19\\
        DCL\cite{c43}&ResNet-50&53.75&45.33&71.41&73.19&42.21&94.73\\
        MaskCOV\cite{c9}&ResNet-50&58.75&46.17&73.57&75.86&50.28&95.82\\
        \midrule
        ViT\cite{c10}&ViT-B&52.50&38.83&53.63&66.95&40.57&96.48\\
        DeiT\cite{c44}&{ViT-B}&54.17&38.67&66.80&69.54&45.34&96.26\\
        TransFG\cite{c28}&{ViT-B}&54.58&40.67&22.38&72.16&21.24&97.14\\

        {SIM-Trans\cite{c54}}&{ViT-B}&{54.58}&{25.00}&{15.46}&{34.76}&{70.69}&{96.70}\\
        Hybrid ViT\cite{c10}&{ViT-B\&ResNet-50}&50.83&37.00&71.74&73.56&18.82&96.48\\
        Mix-ViT\cite{c3}&{ViT-B\&ResNet-50}&60.42&{56.17}&{79.94}&76.30&51.00&{97.36}\\
        {CLE-ViT}\cite{c6}&{Swin-B}&{63.33}&47.17&78.50&{82.14}&{75.21}&\bf 97.58\\
        \midrule
        CSDNet&ResNet-50&61.67&48.17&66.52&78.02&51.05&95.60\\
        {CSDNet}&{ViT-B}&{57.92}&{46.17}&{70.82}&{75.39}&{56.30}&{97.14}\\
        
        CSDNet&Swin-B&\bf 67.92&\bf 60.50&\bf 86.86&\bf 83.17&\bf 76.19&{97.36}\\
        \bottomrule
    \end{tabular}
    \label{tab:resultall}
\end{table*}

\begin{table*}[t]  
    \centering
    \caption{THe classification accuracies of the competing methods on the five subsets of the SoyAgeing dataset. “Avg” denotes the average classification accuracy of the five subsets.
    }
    \begin{tabular}{cccccccc}
        \toprule
         \multirow{2}{*}{Method} &\multirow{2}{*}{Backbone} & \multicolumn{6}{c}{Accuracy (\%)} \\
        \cmidrule(l){3-8}
         &  & R1 & R3 & R4 & R5 &  R6 & Avg \\
        \midrule
        Alexnet\cite{c33}&Alexnet&49.90&44.65&45.15&47.47&37.47&44.93\\
        VGG-16\cite{c34}&VGG-16&72.32&72.53&74.95&71.11&61.31&70.44\\
        ResNet-50\cite{c35}&ResNet-50&70.00&64.24&74.04&72.63&54.85&67.15\\
        \midrule
        SimCLR (FT\cite{c25})&ResNet-50&53.64&45.66&45.35&50.40&35.86&46.18\\
        SimCLR (L)\cite{c25}&ResNet-50&70.00&66.57&64.24&68.38&54.44&64.73\\
        MoCo v2\cite{c37}&ResNet-50&42.93&38.59&38.99&38.99&31.82&38.26\\
        MoCo v2 (L)\cite{c37}&ResNet-50&62.73&56.16&61.31&65.96&49.49&59.13\\
        BYOL (FT)\cite{c38}&ResNet-50&55.35&48.38&50.40&49.60&43.94&49.53\\
        BYOL (L)\cite{c38}&ResNet-50&71.11&66.16&65.76&64.65&56.06&64.75\\
        \midrule
        Cutout (16)\cite{c39}&ResNet-50&70.20&61.92&62.32&69.70&54.24&63.68\\
        Cutout (8)\cite{c39}&ResNet-50&66.87&64.04&67.78&73.43&56.36&65.70\\
        ADL (0.25)\cite{c41}&ResNet-50&53.54&54.34&55.15&52.83&41.92&51.56\\
        ADL (0.5)\cite{c41}&ResNet-50&66.67&58.89&64.75&68.48&49.70&61.70\\
        Cutmix\cite{c42}&ResNet-50&65.56&59.19&64.24&68.79&53.64&62.28\\
        DCL\cite{c43}&ResNet-50&76.87&73.84&76.16&76.16&62.93&73.19\\
        MaskCOV\cite{c9}&ResNet-50&79.80&74.65&79.60&78.28&66.97&75.86\\
        \midrule
        ViT\cite{c10}&{ViT-B}&69.29&64.55&70.40&71.01&59.49&66.95\\
        DeiT\cite{c44}&{ViT-B}&73.03&70.40&69.09&74.65&60.51&69.54\\
        TransFG\cite{c28}&{ViT-B}&74.95&74.55&74.24&76.26&60.81&72.16\\
        {SIM-Trans\cite{c54}}&{ViT-B}&{69.90}&{73.23}&{73.13}&{73.94}&{63.23}&{70.69}\\
        Hybrid ViT\cite{c10}&{ViT-B\&ResNet-50}&77.17&76.97&74.75&76.36&62.53&73.56\\
        Mix-ViT\cite{c3}&{ViT-B\&ResNet-50}&79.29&77.17&77.98&79.19&67.88&76.30\\
        CLE-ViT\cite{c6}&{Swin-B}&{80.81}&{83.33}&{84.24}&\bf 86.36&{75.96}&{82.14}\\
        \midrule
        CSDNet&ResNet-50&81.92&77.47&81.82&79.70&69.19&78.02\\
        CSDNet&{ViT-B}&{75.15}&{76.57}&{77.27}&{78.18}&{69.80}&{75.39}\\
        CSDNet&{Swin-B}&\bf 83.84&\bf 85.15&\bf 85.15&{84.85}&\bf 76.87&\bf 83.17\\
        \bottomrule
    \end{tabular}
    \label{tab:resultR}
\end{table*}

\begin{table}[htp]
    \centering
    \caption{Comparison results on IP102 dataset.
    }
    \begin{tabular}{ccc}
        \toprule
        Method&Backbone&Accuracy (\%)\\
        \midrule
        VGG19\cite{c34}&VGG19&54.1\\
        ResNet-50\cite{c35}&ResNet-50&54.7\\
        API-Net\cite{c47}&ResNet-50&56.9\\
        AP-CNN\cite{c53}&ResNet-50&56.8\\
        ViT\cite{c10}&{ViT-B}&73.4\\
        TransFG\cite{c28}&{ViT-B}&74.8\\
        FFVT\cite{c55}&{ViT-B}&74.0\\
        IELT\cite{c56}&{ViT-B}&74.3\\
        SIM-Trans\cite{c54}&{ViT-B}&70.6\\
        AA-Trans\cite{c48}&{ViT-B}&{75.0}\\
        \midrule
        CSDNet&ResNet-50&71.0\\
        {CSDNet}&{ViT-B}&{74.5}\\
        CSDNet&{Swin-B}&\bf 76.1\\
        \bottomrule
    \end{tabular}
    \label{tab:IP102}
\end{table}

\begin{table}[htp]
    \centering
    \caption{Comparison results on CUB dataset. * indicates that they reproduce their experiments in our environment according to the settings of the original paper.
    }
    \begin{tabular}{ccc}
        \toprule
        Method&Backbone&Accuracy (\%)\\
        \midrule
        DVAN\cite{dvan}&VGG16&79.0\\
        AutoBD\cite{AutoBD}&VGG16&81.6\\
        ResNet-50*\cite{c35}&ResNet-50&82.1\\
        MaskCOV\cite{c9}&ResNet-50&86.6\\
        SPARE\cite{c5}&ResNet-50&86.8\\
        TransFG*\cite{c28}&{ViT-B}&{91.1}\\
        {SIM-Trans*\cite{c54}}&{ViT-B}&{\textbf{91.3}}\\
        IELT*\cite{c56}&{ViT-B}&90.3\\
        DeiT\cite{c44}&{ViT-B}&90.0\\
        ViT\cite{c10}&{ViT-B}&90.6\\
        Mix-ViT*\cite{c3}&{ViT-B}&90.7\\
        \midrule
        CSDNet&ResNet-50&84.1\\
        {CSDNet}&{ViT-B}&{90.9}\\
        CSDNet&{Swin-B}&91.0\\
        \bottomrule
    \end{tabular}
    \label{tab:cub}
\end{table}

\subsubsection{Benchmark Methods}
To ensure a fair comparison, we compare CSDNet with 18 state-of-the-art classification methods on ultra-fine-grained image datasets.
Following the approach in \cite{c3}, we categorize these baselines into 4 groups. 
The first group comprises CNN-based methods. The second focuses on self-supervised learning methods. The third one consists of weakly supervised methods and fine-grained categorization. The fourth group deals with fine-grained and ultra-fine-grained tasks, which rely on visual transformers.

\subsection{Comparison with the State-of-the-Art}
\subsubsection{Evaluation on ultra-fine-grained image datasets} 
We compare the performance of our proposed CSDNet with the state-of-the-art methods on ultra-fine-grained datasets, as shown in Table~\ref{tab:resultall}. 
CSDNet yields significantly superior results compared to other methods, particularly Mix-ViT and CLE-ViT, which are also designed for ultra-fine-grained categorization.
On the Cotton80, SoyLocal, SoyGene, SoyAgeing and SoyGlobal, CSDNet outperforms Mix-ViT by about 7.50$\%$, 4.33$\%$, 6.92$\%$, 6.87$\%$ and 25.19$\%$, and improves about 4.59$\%$, 13.33$\%$, 8.36$\%$, 1.03$\%$ and 0.98$\%$ over CLE-ViT. 

It is worth noting that both SoyGene and SoyGlobal, each having over 1,000 subcategories with limited samples (3 to 11), achieve accuracies of 86.86$\%$ and 76.19$\%$ respectively using CSDNet. 
In addition, the Cotton80 and SoyLocal datasets have an average of only 3 images per subcategory, and with such very limited samples, CSDNet also achieves impressive accuracy rates of 67.92$\%$ and 60.50$\%$, respectively.
We observe a similar trend in SoyAgeing, where the proposed method CSDNet outperforms the second-best method by 1.03$\%$. 
The ultra-fine-grained image SoyAgeing dataset covers five subsets, each containing images collected from a specific cultivation stage.
We report the comparison results of all methods on the five subsets in Table~\ref{tab:resultR}.
The proposed method achieves superior performance on the five datasets compared to other competing methods, proving its effectiveness in the ultra-FGVC task.

\begin{table}[htp]
    \centering
    \caption{{Comparison results on Food101 dataset.
    }}
    \begin{tabular}{ccc}
        \toprule
        Method&Backbone&Accuracy (\%)\\
        \midrule
        ResNet-50\cite{c35}&ResNet-50&83.2\\
        WARN\cite{c96}&WARN-50&85.5\\
        PMG\cite{c45}&ResNet-50&86.9\\
        TransFG\cite{c28}&ViT-B&{92.1}\\
        IELT\cite{c56}&ViT-B&91.4\\
        SIM-Trans\cite{c54}&ViT-B&91.9\\
        µ2Net+\cite{c99}&ViT-L&91.5\\
        \midrule
        CSDNet&ResNet-50&86.5\\
        CSDNet&ViT-B&91.9\\
        CSDNet&Swin-B&\bf 93.5\\
        \bottomrule
    \end{tabular}
    \label{tab:food101}
\end{table}

\subsubsection{Evaluation on fine-grained image datasets}
We compare the results of CSDNet with other advanced methods on fine-grained datasets. 
Table~\ref{tab:resultall} shows that CSDNet achieves strong performance compared with other competitive methods on the crop disease dataset AFD.
To further validate the effectiveness of the proposed method, we evaluate it on a challenging large-scale crop pest dataset IP102. 
Due to the substantial size variability of this dataset, many methods find it challenging to experiment on such a large dataset\cite{c48}. 
The experimental results present in Table~\ref{tab:IP102} demonstrate that the accuracy achieved by CSDNet surpasses the current best performance by 1.1$\%$.
{We also evaluate CSDNet on the widely-used fine-grained image dataset, CUB-200-2011, and Food101 in Table~~\ref{tab:cub} and~\ref{tab:food101}.
}
These experimental results on fine-grained datasets demonstrate the effectiveness and generalization ability of CSDNet in fine-grained visual categorization.

Furthermore, to assess the generalizability of CSDNet, we test it across various network architectures. We use the CNN network ResNet-50\cite{c35} as the backbone for the experiments. {For a fair comparison, we also experimented on using ViT-B\cite{c10} as the backbone.}
The experimental results in Table~\ref{tab:resultall} and ~\ref{tab:resultR} show that CSDNet using ResNet-50 outperforms almost all CNN methods on ultra-fine-grained tasks, and also has competitive performance when compared with the visual transformer methods. 
Besides, jointly considering Table~\ref{tab:resultall}, ~\ref{tab:resultR}, ~\ref{tab:IP102}, ~\ref{tab:cub}, and ~\ref{tab:food101}, CSDNet using ResNet-50 achieves strong performance on both ultra-fine-grained tasks and fine-grained tasks, demonstrating that CSDNet can be combined with different backbones and achieve competitive performance.

\subsubsection{Visualization}
\begin{figure*}[htp]
  \centering
  \includegraphics[width=0.9\textwidth]{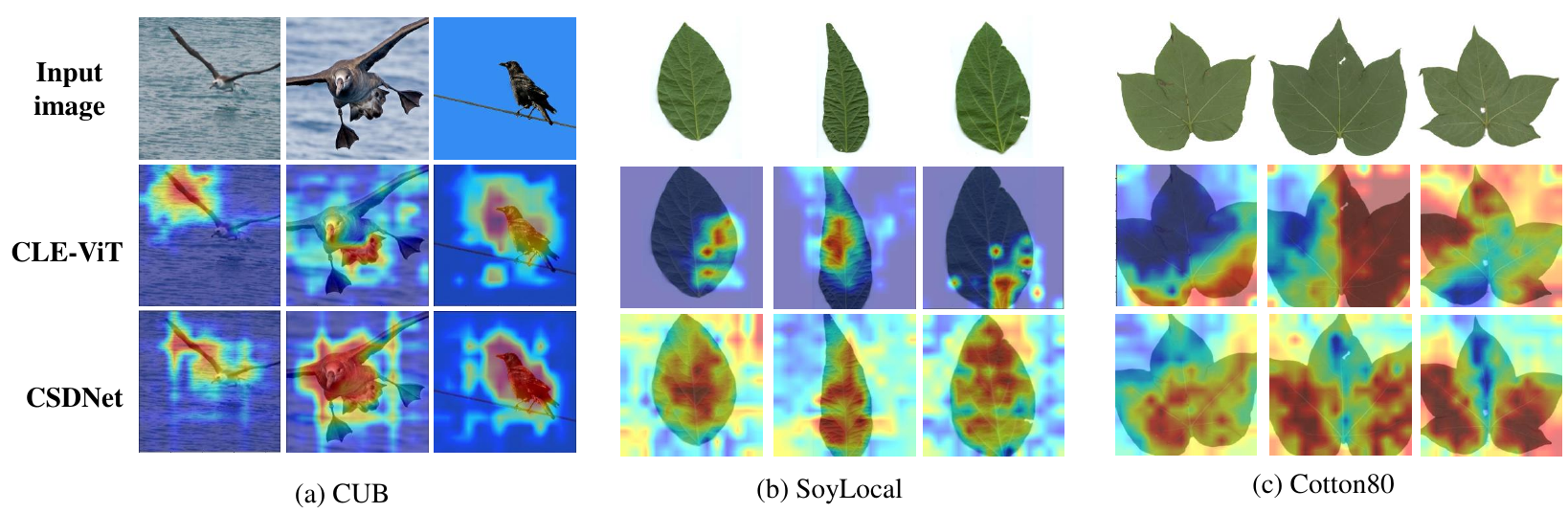} 
  \caption{Visual comparison of, CLE-ViT (line 2) and proposed CSDNet (line 3) with the original image (line 1) on the CUB, SoyLocal, Cotton80 dataset.}
  \label{fig:camcub}
\end{figure*}

With the question of what information in the image enables the network to classify it visually, we present the CSDNet visualization results as shown in Fig.\ref{fig:camcub}. 
We visualized the results of CLE-ViT and the proposed CSDNet on the CUB, SoyLocal, and Cotton80 datasets. 
In these visualizations, the highlighted red areas indicate regions that make significant contributions to visual categorization. 
We observe that CLE-ViT has limitations in its ability to concentrate on discrepancy areas, and it may even primarily focus on background regions.
In contrast, CSDNet is able to focus on more comprehensive regions, specifically capturing more detailed discrepancies specific to subcategories, such as bird head, leaf tips, and leaf veins.

\begin{figure*}[htp]
  \centering
  \includegraphics[width=0.9\textwidth]{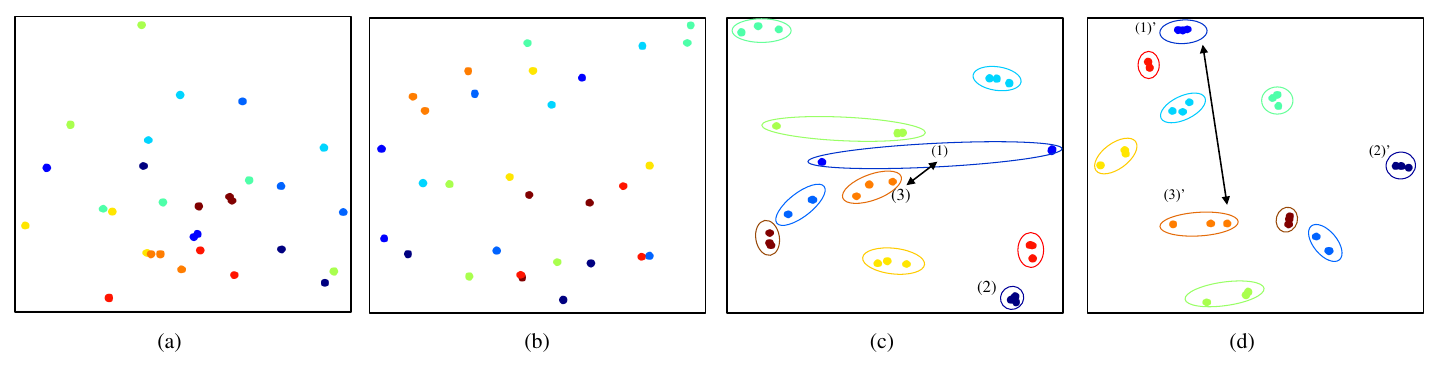} 
  \caption{Visualize the tSNE visualization of features learned on the Cotton80 dataset for four settings: (a) Baseline, (b) Baseline + SSDP, (c) Baseline + SSDP + DDL, and (d) Baseline + SSDP + DDL + SSDT (full model). Each color represents a unique subcategory (a total of 10 subcategories).}
  \label{fig:tsne}
\end{figure*}

To provide a more intuitive demonstration of the importance of mining and transferring subcategory-specific discrepancies, we apply tSNE visualizations\cite{c88} to display the feature distribution after sequentially stacking the modules we proposed in the paper, as illustrated in Fig. \ref{fig:tsne}. 
The samples are randomly selected from 10 classes in the test set. It is worth mentioning that since Cotton80 only has 3 images per subcategory, there are only 3 scatter points for each subcategory.
{We observe that in the baseline, as shown in Fig.~\ref{fig:tsne} (a), many sample points are concentrated and even overlap, which is unfavorable for classification. }
{When SSDP module~(Section.~\ref{sec3.c}) is added and enhanced images are generated, the knowledge of subcategory-specific differences is reinforced, causing the sample points to disperse and preventing the model from overfitting, as shown in (b).}
{After introducing DDL~(Section.~\ref{sec3.d}), we can observe from (c) that the samples within each subcategory are drawn closer together, creating clearer boundaries between subcategories.}
{Finally, by adding SSDT~(Section.~\ref{sec3.e}), originally overlapping subcategories are separated, as seen with subcategory (2) in (c) compared to (2)’ in (d).
At the same time, it brings closer the samples of the same subcategory that are far apart, as seen with subcategory (1) in (c) compared to (1) in (d). The distribution between subcategories is more distinct, as observed with the distance between subcategories (1) and (3) in (c) compared to (1) and (3)’ in (d).
}

%

\begin{figure}
  \centering
  \includegraphics[width=0.9\linewidth]{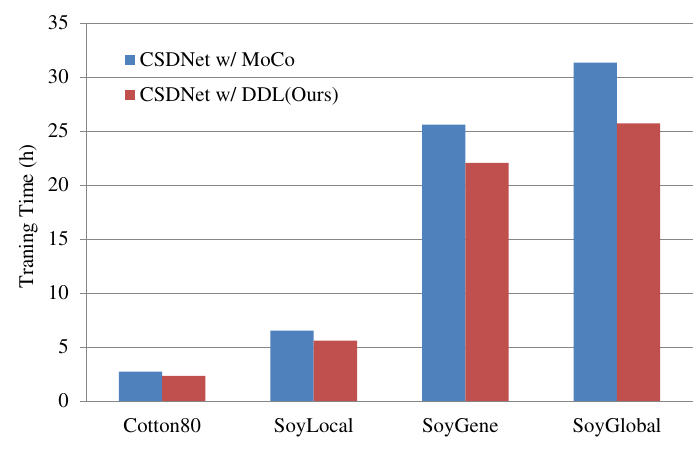}
  \caption{Comparison of CSDNet w/ DDL and CSDNet w/ MoCo training duration.}
  \label{fig:timemoco}
\end{figure}

\begin{figure*}[htp]
  \centering
  \includegraphics[width=0.9\textwidth]{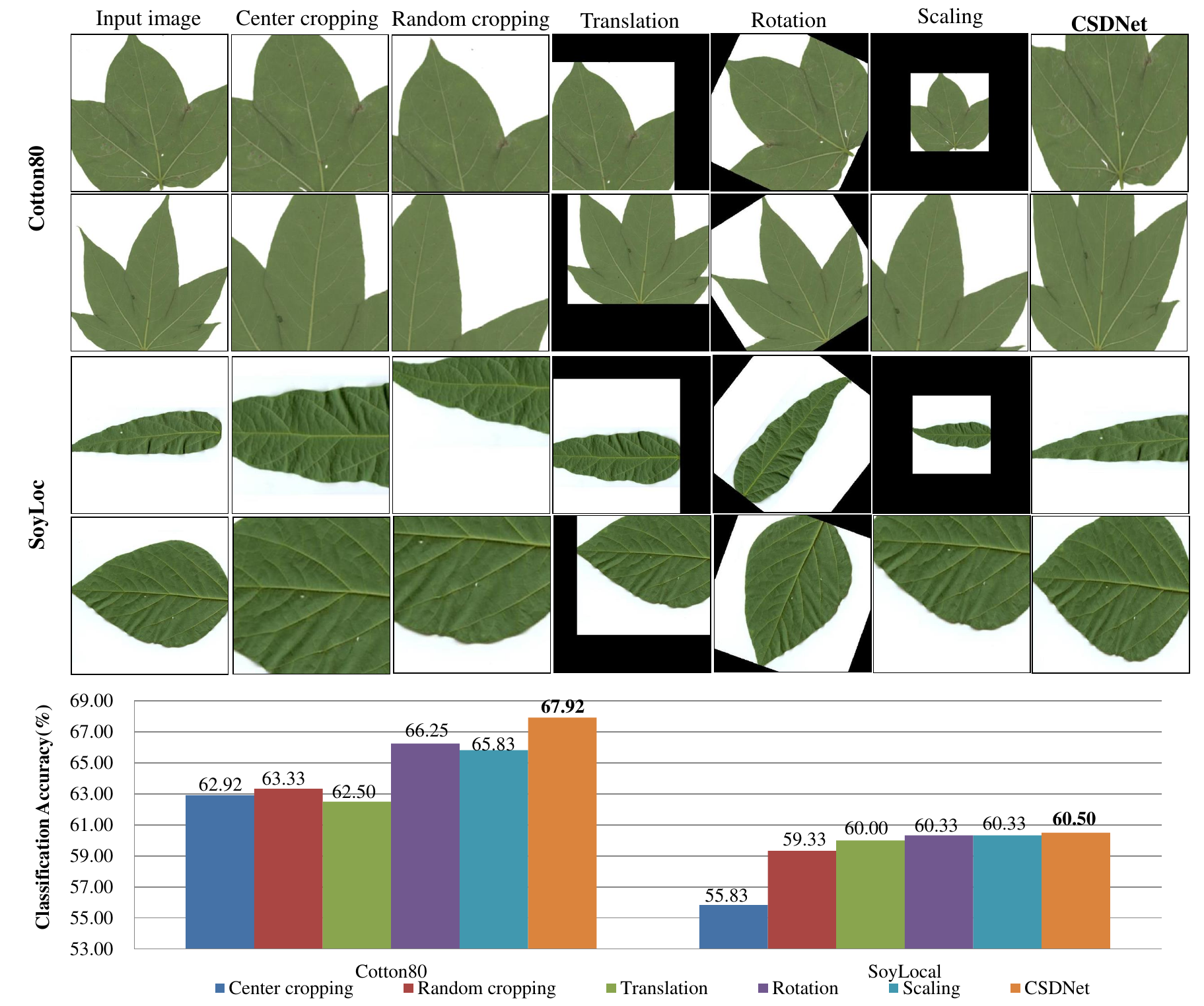}
  \caption{{Comparison of SSDP data augmentation with other traditional data augmentation methods. From left to right, the sequence is the augmentation effect of the input image, center cropping, random cropping, rotation scaling, translation, and CSDNet. The bar chart represents the classification accuracy obtained using different data augmentation methods.}}
  \label{fig:augcomparation}
\end{figure*}

\begin{table} 
    \centering
    \caption{Ablation studies of the proposed components. SSDP, DDL, and SSDT denote the proposed Subcategory-Specific Discrepancy Parsing, Dynamic Discrepancy Learning, and Subcategory-Specific Discrepancy Transfer, respectively.}
    \resizebox{0.48\textwidth}{!}{
    \begin{tabular}{ccccccc}
        \toprule
        \multirow{2}{*}{SSDP}&\multirow{2}{*}{DDL}&\multirow{2}{*}{SSDT}&\multicolumn{4}{c}{Accuracy (\%)} \\
        \cmidrule(l){4-7}
        &&& Cotton80 & SoyLocal & SoyGene &  SoyGlobal  \\
        \midrule
        -&-&-&61.25&51.83&81.77&69.83\\
        \checkmark&-&-&61.67&53.83&83.93&70.06\\
        \checkmark&\checkmark&-&63.75&54.33&86.75&72.96\\
        \checkmark&-&\checkmark&66.25&57.67&84.24&74.79\\
        \midrule
        \checkmark&\checkmark&\checkmark&\bf 67.92&\bf 60.50&\bf 86.86&\bf 76.19\\
        \bottomrule
    \end{tabular}
    }
    \label{tab:Ablation_module}
\end{table}

\subsection{Ablation Studies}
In order to demonstrate the effectiveness of CSDNet, we perform comprehensive ablation studies on 4 datasets~(Cotton80, SoyLocal, SoyGene, and SoyGlobal) to analyze how key components or hyperparameters within our method impact its performance.

\subsubsection{Effectiveness of the proposal key components}
In this section, we verify the Effectiveness of the proposed key components proposed by CSDNet, \emph{i.e.}, the Subcategory-Specific Discrepancy Parsing module~(SSDP, Section.~\ref{sec3.c}), the Dynamic Discrepancy Learning module~(DDL, Section.~\ref{sec3.d}) and the Subcategory-Specific Discrepancy Transfer module~(SSDT, Section.~\ref{sec3.e}).
In Table~\ref{tab:Ablation_module}, the first row shows the performance of a baseline model trained using the same procedure as the final model. It is obvious that the classification results exhibit steady improvement as we incorporate our proposed components.
The results in the second and third lines show that expanding samples at the instance level and feature level helps mitigate the impact of limited samples.
In the fourth row, the results show that the SSDT module learns more subcategory-specific discrepancies from raw and augmented samples by performing self-distillation, thereby improving the generalization ability of the model.
Besides, the best results were achieved when combining the SSDT module and the DDL module, indicating they are complementary to each other.

\begin{table}[htp]  
    \centering
    \caption{Comparison of performance and efficiency with~(without) SSDT.
    FPS means frames per second, which is the number of images processed per second.}
    \resizebox{0.5\textwidth}{!}{
    \begin{tabular}{cccccc}
        \toprule
         \multirow{2}{*}{Setting}& \multicolumn{4}{c}{Accuracy (\%)}& \multirow{2}{*}{FPS} \\
        \cmidrule(l){2-5}
         & Cotton80 & SoyLocal & SoyGene &  SoyGlobal &  \\
        \midrule
        CSDNet w/o SSDT &63.75&54.33&86.75&72.96 & 42.13\\
        CSDNet w/ SSDT &\textbf{67.92} &\textbf{60.50} &\textbf{86.86} &\textbf{76.19} & \textbf{113.67}\\
        \bottomrule
    \end{tabular}
    }
    \label{tab:embedding}
\end{table}

\begin{table}[htp]  
    \centering
    \caption{Comparison of different measurement learning methods.}
    \resizebox{0.5\textwidth}{!}{
    \begin{tabular}{cccccc}
        \toprule
        \multirow{2}{*}{Loss}& \multicolumn{4}{c}{Accuracy (\%)} \\
        \cmidrule(l){2-5}
         & Cotton80 & SoyLocal & SoyGene &  SoyGlobal\\
        \midrule
        Multi-similarity loss\cite{c91} &67.50 &60.50 &87.19 &74.75 \\
        Margin loss\cite{c100} &66.25 &60.17 &\textbf{87.23} &75.97 \\
        Triplet loss\cite{c92} &65.83 &59.17 &87.19 &76.06 \\
        \midrule
        SSDP &\textbf{67.92} &\textbf{60.50} &86.86 &\textbf{76.19}\\
        \bottomrule
    \end{tabular}
    }
    \label{tab:conloss}
\end{table}

\begin{table}[htp]
    \centering
    \caption{{Comparison of CSDNet w/ DDL and CSDNet w/ MoCo.}}
    \resizebox{0.5\textwidth}{!}{
    \begin{tabular}{cccccc}
        \toprule
        \multirow{2}{*}{Memory Method}& \multicolumn{4}{c}{Accuracy (\%)} \\
        \cmidrule(l){2-5}
         & Cotton80 & SoyLocal & SoyGene &  SoyGlobal\\
        \midrule
        CSDNet w/ MoCo\cite{c7} &67.50 &60.00 &\textbf{87.35} &73.46 \\
        CSDNet w/ DDL (ours) &\textbf{67.92} &\textbf{60.50} &86.86 &\textbf{76.19}\\
        \bottomrule
    \end{tabular}
    }
    \label{tab:memoryqueue}
\end{table}

\begin{table}[htp]
    \centering
    \caption{{The accuracy(\%) of scaling the image size with Center cropping, Random cropping, and CSDNet. We set up three scales, 244, 448, and 672. 224 means 224$\times$244}. }
    \resizebox{0.5\textwidth}{!}{
    \begin{tabular}{cccccccccc}
        \toprule
         \multirow{3}{*}{DataSet}& \multicolumn{3}{c}{Center Cropping}& \multicolumn{3}{c}{Random Cropping}& \multicolumn{3}{c}{CSDNet} \\
        \cmidrule(lr){2-4} \cmidrule(lr){5-7}\cmidrule(lr){8-10}
         & 224 & 448 & 672& 224 & 448 & 672 & 224 & 448 & 672 \\
        \midrule
        Cotton80&59.17&62.92&67.50&60.83&63.33&67.92&\textbf{62.50}&\textbf{67.92}&\textbf{68.33} \\
        SoyLocal &50.33&55.83&59.33&\textbf{56.83}&59.33&60.17&54.33&\textbf{60.50}&\textbf{60.67}\\
        \bottomrule
    \end{tabular}
    }
    \label{tab:size}
\end{table}

\subsubsection{Importance of the Subcategory-Specific Discrepancy Parsing}
{We investigate the influence of our SSDP augmentation method through experiments with traditional augmentation methods, including center cropping, random cropping, translation, rotation, and scaling. We present the visualization results of data augmentation and the accuracy of different augmentation methods in Fig.~\ref{fig:augcomparation}. SSDP calculates a mask based on the image features and augments the image by filtering the background redundant information through the mask without destroying the leaf structure as much as possible. Compared with the above five traditional augmentation strategies, CSDNet employs an adaptive augmentation strategy to achieve more accurate results. In particular, CSDNet has improved by 4.59\% and 5.00\% on the Cotton80 dataset compared to center cropping and random cropping, respectively.}
{We also compared the accuracy at different image scales, including 224$\times$224, 448$\times$448, and 672$\times$672 in Table~\ref{tab:size}, using center cropping, random cropping, and CSDNet as augmentation strategies. From the results, it can be observed that as the image scale increases, the accuracy of all augmentation methods improves.
It is noteworthy that CSDNet always outperforms other methods at the same scale. However, given that larger image scales lead to longer training times, we choose 448$\times$448 as the final image scale.}


\subsubsection{Effectiveness of the contrastive loss}The comparisons of the performance with different measurement loss are shown in Table~\ref{tab:conloss} to verify the effectiveness of contrastive loss.
The contrastive loss outperforms both the Multi-similarity loss~\cite{c91} and the Triplet loss~\cite{c92}.
Although the contrastive loss exhibits a slight disadvantage compared to the Margin loss~\cite{c100} on the SoyGene dataset, it performs better than the Margin loss overall.


\subsubsection{Advantages of memory queue}To deeply investigate the effectiveness of our memory queue, we add a comparison experiment with MoCo~\cite{c7} that builds a dynamic dictionary with a queue and a moving-averaged encoder called momentum encoder to replace the memory queue. 
%
The experimental results are presented in Table \ref{tab:memoryqueue}. Our method demonstrates better performance across most datasets, with a slightly lower performance observed on SoyGene.
%
However, since MoCo relies on the momentum encoder to encode historical features, this leads to an improvement in training time of about 16.07\% to 21.91\% as shown in Fig. \ref{fig:timemoco}.


\subsubsection{Importance of the Subcategory-Specific Discrepancy Transfer} 
{
To demonstrate the importance of the proposed Subcategory Specific Discrepancy Transfer~(SSDT) module mentioned in Section~\ref{sec3.e}. We summarize the performance of CSDNet with~(without) SSDT on multiple datasets and the inference speed of an Nvidia RTX3090 in Table~\ref{tab:embedding}. 
From Table~\ref{tab:embedding}, we find that after adding the SSDT to CSDNet, there is a performance improvement and the inference time has been greatly reduced. 
The reason is that SSDT helps the CSDNet learn more knowledge about subcategory discrepancies and enhances CSDNet's ability to discriminate subcategories, so that CSDNet can make accurate predictions using only raw samples, greatly reducing time-consuming and memory-demanding.
}
%

\subsection{Hyperparameter Analyses}
\begin{figure}
  \centering
  \includegraphics[width=0.9\linewidth]{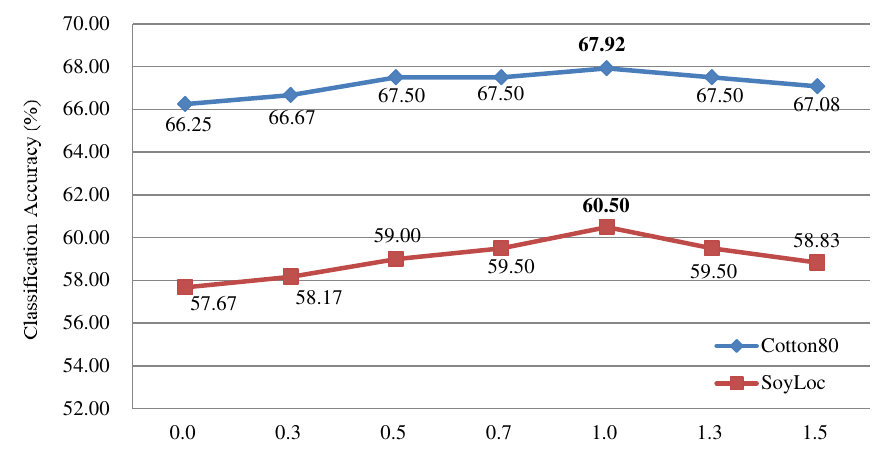}
  \caption{Ablation study on the weight $\alpha$ of $\mathcal L_{qc}$ on Cotton80 and SoyLocal dataset.}
  \label{fig:lossa}
\end{figure}

The performance comparison of different values of $\alpha$, representing the weight of loss $\mathcal L_{qc}$ in Eq.(\ref{eq:l}) is shown in Fig.~\ref{fig:lossa}.
According to the figure, we find that the performance gradually increases until the value of $\alpha$ exceeds 1.0.
This performance drop may be attributed to the fact that as the weight $\alpha$ increases, the model may prioritize feature learning over the ultra-fine-grained categorization, resulting in degraded classification results.
Consequently, we have chosen to set $\alpha$ to 1.0 for all datasets.

\subsubsection{Role of the loss weight $\beta$}
\begin{figure}
  \centering
  \includegraphics[width=0.9\linewidth]{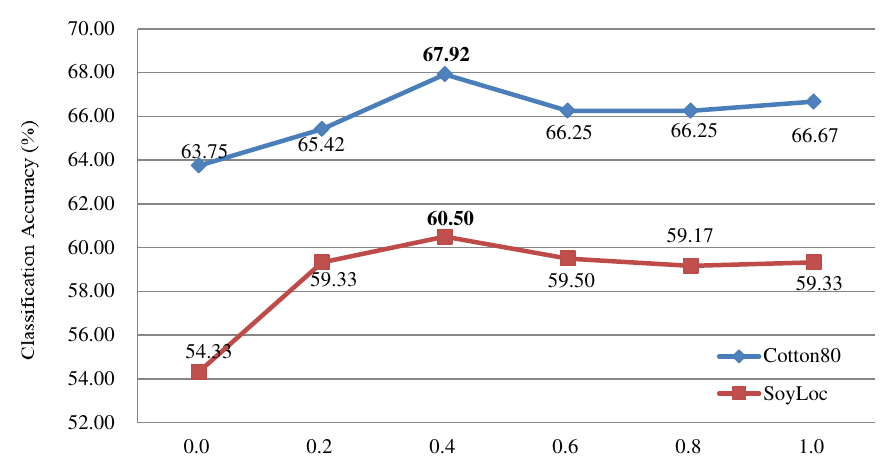} 
  \caption{Ablation study on the weight $\beta$ of $\mathcal L_{ssdt}$ on Cotton80 and SoyLocal dataset.}
  \label{fig:lossb}
\end{figure}

To accurately transfer subcategory-specific discrepancies through $\mathcal L_{ssdt}$ in Eq.(\ref{eq:l}), what should be the weight $\beta$ of the $\mathcal L_{ssdt}$? 
We conduct experiments to analyze the value of $\beta$ and present the results in Fig.~\ref{fig:lossb}.
It can be clearly observed that the best performance is reached when the $\beta$ value reaches 0.4 and decreases after exceeding 0.4.
The reason may be that if the value of $\beta$ is too large, the Subcategory-Specific Discrepancy Transfer module would completely dominate training and ignore the learning of discriminative features, resulting in sub-optimization and performance degradation.
Therefore, we set $\beta$ to 0.4 for all datasets.

\subsubsection{The length of Memory Queue}
In a supervised learning scenario, model parameters are updated rapidly, resulting in significant feature differences between distant batches, which is not conducive to contrastive learning.
Therefore, we conduct comparison experiments to explore how long the memory queue in Eq.(\ref{mqueue}) should be stored.
Table~\ref{tab:queue} presents a summary of performance for different memory queue lengths, $\iota$, ranging from 0 to 3.
From the results, it is evident that the performance with $\iota>0$ is superior to that with $\iota=0$, indicating the benefits of our dynamic contrastive learning approach over traditional contrastive learning. Note that $\iota = 0$ implies that dynamic contrastive learning is equivalent to the traditional approach.
In addition, we observe that Cotton80 and SoyLocal, SoyGene, and SoyGlobal, have different optimal queue lengths $\iota$.
This can be attributed to the difference in the number of subcategories in the dataset. The greater the number of subcategories, the greater the model parameter updates during the training process, resulting in greater differences between features of different batches.
Accordingly, we set $\iota$ to 2 for datasets with less than 1000 subcategories and 1 for the others.

\subsubsection{The binarization threshold $\lambda$}
\begin{figure}
  \centering
  \includegraphics[width=0.9\linewidth]{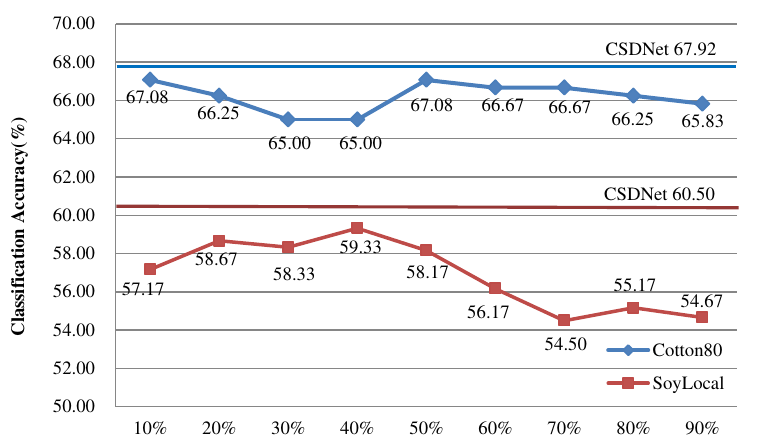} 
  \caption{{Ablation study on the threshold $\lambda$. Two horizontal lines represent the accuracy of our CSDNet on Cotton80 and SoyLoc, others represent the accuracy corresponding to different fixed thresholds.}}
  \label{fig:threshold}
\end{figure}

{The calculation of the mean value of each image's features as a threshold is an adaptive enhancement strategy. Since the proportion of the foreground to the background in each image varies, choosing a fixed threshold is unreasonable. To verify the effectiveness of the adaptive threshold, we set the threshold $\lambda$ to range from 10\% to 90\% of the current maximum activation value and conducted experiments, as shown in Fig. \ref{fig:threshold}. The horizontal line represents the results of our enhancement method, while the line graph shows the results of experiments with a fixed threshold. The experimental results indicate that our adaptive enhancement strategy achieved the best accuracy. }


\begin{table}[t]
    \centering
    
    \caption{Analyses of hyperparameter $\iota$ on Cotton80, SoyLocal, SoyGene and SoyGlobal datasets.}
    \begin{tabular}{ccccc} 
        \toprule 
        \multirow{2}{*}{$\iota$}& \multicolumn{4}{c}{Accuracy (\%)} \\
        \cmidrule(l){2-5}
         & Cotton80 & SoyLocal & SoyGene & SoyGlobal \\
        \midrule 
        0  & 67.08  & 59.00  & 85.80  & 74.18  \\
        1  & 67.28  & 59.83  & \textbf{86.86}  & \textbf{76.19}  \\
        2  & \textbf{67.92}  & \textbf{60.50}  & 86.33  & 75.70  \\
        3  & 66.67  & 59.83  & 86.54  & 74.03  \\
        \bottomrule 
    \end{tabular}
    \label{tab:queue}
\end{table}

%% file: sections/conclusion.tex
\section{Conclusions}\label{sec:five}
This paper presents a novel framework, CSDNet, to tackle the challenge of ultra-fine-grained visual categorization of limited samples through the utilization of contrastive learning and self-distillation.
To address the challenges posed by limited samples, we introduce two key modules: the Subcategory-Specific Discrepancy Parsing module, which generates adaptive augmented samples by parsing subcategory-specific discrepancies, and the Dynamic Discrepancy Learning module, which employs a dynamic memory queue to store historical features and integrates them into the contrastive learning process. These modules enhance sample diversity at the instance and feature levels respectively.
Furthermore, we introduce the Subcategory-Specific Discrepancy Transfer module, which extracts more subcategory-specific discrepancies knowledge from limited samples through logit prediction level distillation between raw samples and augmented samples without the need for additional data annotation.
Experiments demonstrate that CSDNet has highly competitive performance on eight public datasets, excelling in ultra-fine-grained visual categorization and fine-grained visual categorization.